# A Publicly Available Cross-Platform Lemmatizer for Bulgarian

**Grigor Iliev, Nadezhda Borisova, Elena Karashtranova, Dafina Kostadinova**

*South-West University, Blagoevgrad, Bulgaria*

**Abstract:** *Our dictionary-based lemmatizer for the Bulgarian language presented here is distributed as free software, publicly available to download and use under the GPL v3 license[1]. The presented software is written entirely in Java and is distributed as a GATE plugin. To our best knowledge, at the time of writing this article, there are not any other free lemmatization tools specifically targeting the Bulgarian language. The presented lemmatizer is a work in progress and currently yields an accuracy of about 95% in comparison to the manually annotated corpus BulTreeBank-Morph[2], which contains 273933 tokens [1].*

## 1. INTRODUCTION

Lemmatization is a fundamental natural language processing (NLP) task which automates the process of word normalization. The correct identification of the normalized form of a word is of particular importance for several other NLP tasks and can improve the accuracy of tasks like information extraction and information retrieval. The implementation of a lemmatizer can be a challenging task, especially in cases of highly inflectional languages such as the Bulgarian language. For example, the Bulgarian adjective 'рядък' *rădăk* - 'rare') has 9 inflected word forms (Table 1).

Table 1. The inflected word forms of the adjective 'рядък' (*rădăk* - 'rare')

| Cyrillic script | Latin scirpt (ISO 9) | Gender | Number | Article | Extended form |
|---|---|---|---|---|---|
| редкия | redkiâ | masculine | singular | definite | no |
| редкият | redkiât | masculine | singular | definite full | no |
| рядка | râdka | feminine | singular | indefinite | no |
| рядката | râdkata | feminine | singular | definite | no |

---

[1] https://github.com/grigoriliev/BGLangTools/releases
[2] http://www.bultreebank.org/btbmorf/

| рядко | râdko | neuter | singular | indefinite | no |
| рядкото | râdkoto | neuter | singular | definite | no |
| редки | redki | – | plural | indefinite | no |
| редките | redkite | – | plural | definite | no |
| редки | redki | masculine | singular | – | yes |

## 2. IMPLEMENTATION

The BGLemmatizer package is part of the BGLangTools project. For the implementation of the BGLemmatizer, we decided to reuse freely available lexical resources. We focused on two projects - BG Office[1] and Bulgarian-language Wiktionary[2]. We implemented support in BGLangTools for automatic retrieval of all lemmas provided by these projects and automatic word form generation for the lemmas having inflected forms.

The word forms of lexemes are generated by means of inflectional paradigms. In the Bulgarian language there are 187 inflectional paradigms (types), which cover almost all words with inflected forms [2]. To inflect all word forms of a lemma, we only need to know the inflectional paradigm to which the lemma belongs. In a nutshell, we need to know the lemma type. For example, to retrieve a list of all word forms of the Bulgarian adjective 'рядък' (<i>râdăk</i> - 'rare'), which is of type "83", we will use the following code:

```
WordEntry[] forms;
forms = BgWordFormGenerator.generateWordForms("рядък", "83");
```

Since BG Office and Bulgarian-language Wiktionary projects are work in progress and it is expected that the provided lexical resources will grow in time, BGLangTools provides an option to scan the projects data and import the newly added lemmas in its database. Currently, the dictionary contains 65 376 lemmas and 1 017 595 inflected forms. The number of inflected forms will grow, because at the time this paper was written the inflection of verbs has been partially implemented - only the most frequently used word forms could be generated. Note that in the Bulgarian language a verb can have up to 2000 word forms [3].

Most of the currently available part-of-speech (POS) taggers for Bulgarian use the BulTreeBank tagset (BTB-TS) scheme [4] to annotate words with POS tags. For efficiency reasons, in BGLangTools the grammatical information about a word entry is stored in a 32-bit integer, but for interoperability purposes, we implemented support for converting a BTB-TS POS tag to a 32-bit integer value and vice versa.

---

[1] http://bgoffice.sourceforge.net
[2] http://bg.wiktionary.org

### 2.1. Retrieving a list of lemmas from the BG Office project

The goal of the BG Office project is to add spell check support for Bulgarian in various open source projects like OpenOffice (openoffice.org) and Mozilla (mozilla.org). BG Office provides software packages for developers[1], which contain lexical resources for Bulgarian, where the lemmas are sorted according to inflectional type and stored in separate files.

In the latest version (4.1) of the developers package the lexical resources are located in the 'data' directory. In this directory and its sub-directories, there are several files with .dat extension whose names start with 'bg', followed by a number and an optional one-letter suffix. The number with the optional suffix designates the inflectional type of the lemmas contained in the file.

BGLangTools provides support for retrieval of all lemmas from a BG Office developers package. For every lemma the corresponding inflectional forms are generated and added to the dictionary. This can be done by means of the following code:

```
String bgofficePath = "/packages/bgoffice-4.1/";
BgDictionary dict = new BgDictionary();
BgOfficeScanner.getInstance().scan(bgofficePath, dict);
```

The bgofficePath variable specifies the absolute path to the top level directory of the BG Office package.

### 2.2. Retrieving a list of lemmas from the Bulgarian-language Wiktionary project

The goal of the Wiktionary project is to provide a complete international dictionary under a free license. The backup dumps of the content are available for free download[2].

The support for processing the Bulgarian-language Wiktionary dump file[3] is implemented in BGLangTools by using the Bliki engine[4]. The following code can be used to process the specified dump file, retrieve all lemmas, generate the corresponding inflectional forms and import them to the specified dictionary:

---

[1] http://sourceforge.net/projects/bgoffice/files/For%20Developers/
[2] http://dumps.wikimedia.org/
[3] http://dumps.wikimedia.org/bgwiktionary/latest/bgwiktionary-latest-pages-articles.xml.bz2
[4] https://code.google.com/p/gwtwiki/

```
String dump;
dump = "/dumps/bgwiktionary-latest-pages-articles.xml.bz2";
BgDictionary dict = new BgDictionary();
BgWiktionaryScanner.getInstance().scan(dump, dict);
```

The `dump` variable specifies the absolute path to the Wiktionary dump file.

### 2.3. Using the built-in dictionary

At the time this article was written, BGLangTools kept in internal format 1 082 971 tokens, retrieved from the BG Office project and the Bulgarian-language Wiktionary project. In order to load the internally stored tokens with their grammatical information, one can use the following code:

```
BgDictionary dict = BGLangTools.loadBuiltinDictionary();
```

## 3. USING THE BGLEMMATIZER PLUGIN FOR GATE

GATE (General Architecture for Text Engineering) is a mature language processing framework capable of handling various text processing tasks[1]. It is an open source project with over 15 years of development. We implemented support for running BGLemmatizer as GATE plugin.

Part-of-speech information is required for every token so that the BGLemmatizer plugin can work properly. This information can be automatically obtained by POS taggers like the LingPipe POS tagger, which is also available as GATE plugin. An example configuration of the BGLemmatizer plugin is shown in Figure 1, where the plugin is configured to retrieve the POS tag of a token from the 'category' feature of the 'Token' annotation and to set the resulting lemma as a 'lemma' feature in the same annotation.

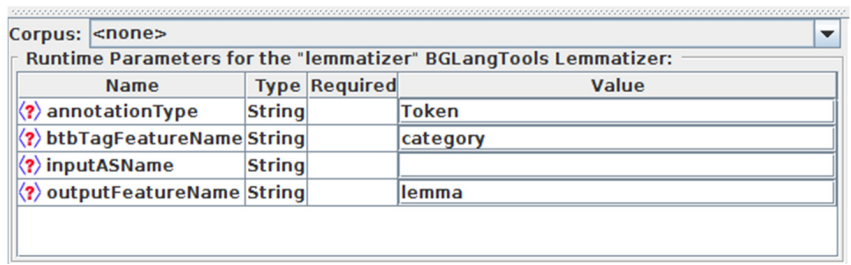

Figure 1. Runtime parameters of the BGLemmatizer plugin in GATE

---

[1] http://gate.ac.uk/

## 4. FUTURE WORK

We plan to integrate the BGLemmatizer with other linguistic software tools. One of our aims is to achieve better results and accuracy of part-of-speech annotation of Bulgarian texts with high levels of grammatical errors, which is the case with texts written by people learning Bulgarian as a foreign language. Further, we plan to extend BGLangTools with APIs, which will be useful in the fields of automatic text generation and information extraction.

## 5. ACKNOWLEDGEMENTS

This research was funded by the South-West University "Neofit Rilski" grant SRP-B4/15.